\documentclass[12pt]{spieman}  
\usepackage{amsmath,amsfonts,amssymb}
\usepackage{graphicx}
\usepackage{booktabs}
\usepackage{setspace}
\usepackage{tocloft}
\usepackage{lineno}
\usepackage[color]{changebar}  
\cbcolor{white}
\usepackage[dvipsnames,table]{xcolor}

\newcommand{\TOne}{T\ensuremath{_1}}
\newcommand{\TTwo}{T\ensuremath{_2}}

\usepackage{pifont}
\newcommand{\cmark}{\textcolor{green!70!black}{\ding{51}}}
\newcommand{\xmark}{\textcolor{red}{\ding{55}}}  

\title{ECLARE: Efficient cross-planar learning for anisotropic resolution enhancement}

\author[a,*]{Samuel~W.~Remedios}
\author[b]{Shuwen Wei} 
\author[b]{Shuo Han}
\author[b]{Jinwei Zhang}
\author[b]{Aaron Carass}
\author[c,d]{Kurt~G. Schilling} 
\author[e]{Dzung~L. Pham}
\author[b]{Jerry~L. Prince}
\author[f]{Blake~E. Dewey}

\affil[a]{Department of Computer Science, Johns Hopkins University, Baltimore MD, USA}
\affil[b]{Image Analysis and Communications Laboratory, Department of Electrical and Computer Engineering, Johns Hopkins University, Baltimore MD, USA}
\affil[c]{Department of Radiology and Radiological Sciences, Vanderbilt University Medical Center, Nashville TN, USA}
\affil[d]{Vanderbilt University Institute of Imaging Science, Vanderbilt University Medical Center, Nashville TN, USA}
\affil[e]{Department of Radiology, Uniformed Services University, Bethesda MD, USA}
\affil[f]{Department of Neurology, Johns Hopkins School of Medicine, Baltimore MD, USA}

\cftpagenumbersoff{figure}
\cftpagenumbersoff{table} 
\begin{document} 
\maketitle

\begin{abstract}\\
\noindent\textbf{Purpose:}\quad In clinical imaging, magnetic resonance~(MR) image volumes are often acquired as stacks of 2D slices with decreased scan times, improved signal-to-noise ratio, and image contrasts unique to 2D MR pulse sequences.
While this is sufficient for clinical evaluation, automated algorithms designed for 3D analysis perform poorly on multi-slice 2D MR volumes, especially those with thick slices and gaps between slices.
Super-resolution (SR) methods aim to address this problem, but previous methods do not address all of the following: slice profile shape estimation, slice gap, domain shift, and non-integer or arbitrary upsampling factors. 

\noindent\textbf{Approach:}\quad In this paper, we propose ECLARE (Efficient Cross-planar Learning for Anisotropic Resolution Enhancement), a self-SR method that addresses each of these factors.  
ECLARE uses a slice profile estimated from the multi-slice 2D MR volume, trains a network to learn the mapping from low-resolution to high-resolution in-plane patches from the same volume, performs SR with anti-aliasing, and respects the image FOV during resampling.
We compared ECLARE to cubic B-spline interpolation, SMORE, and other contemporary SR methods. 
We used realistic and representative simulations on human head MR volumes so that quantitative performance against ground truth can be computed.
Specifically, healthy $\TOne$-w and people with MS $\TTwo$-w FLAIR datasets were used for evaluations.
We used the peak signal to noise ratio and structural similarity index measure as signal recovery metrics.
We additionally used two independent brain parcellation algorithms, SLANT and SynthSeg, to compute the consistency Dice similarity coefficient and the $R^2$ coefficient of determination, respectively, as comparison metics.

\noindent\textbf{Results:}\quad For images with up to 5 mm of slice thickness and up to 1.5 mm of gap, ECLARE achieves greater mean PSNR and SSIM compared to other methods.
In representative regions of interest, such as the ventricles, caudate, cerebral white matter, and cerebellar white matter, ECLARE performs comparably or better than other approaches.
These trends are similar for both investigated datasets.

\noindent\textbf{Conclusions:}\quad The use of slice profile estimation, FOV-aware resampling, and self-SR allowed ECLARE to robustly super-resolve anisotropic images without the need of external training data.
Future work will investigate the utility of ECLARE on other organs, species,  modalities, and resolutions.
Our code is open-source and available at \href{https://www.github.com/sremedios/eclare}{https://www.github.com/sremedios/eclare}.
\end{abstract}

\keywords{inverse problem, magnetic resonance imaging, slice gap, super-resolution}

{\noindent \footnotesize\textbf{*}S.W.R.~\linkable{samuel.remedios@jhu.edu} }


\section{Introduction}
\label{sec:introduction}
Magnetic resonance~(MR) imaging~(MRI) is an essential clinical tool in diagnosis, monitoring, management, and research across many health conditions.
Multi-slice 2D MRI uses slice selection~\cite{brown2014magnetic} to acquire multiple evenly spaced 2D slices, which are stacked to create a 3D volume.
In contrast, 3D MRI uses two phase encode directions to acquire 3D volumes directly~\cite{elmaouglu2011mri}. 
Although 3D MRI has many advantages, including greater utility in downstream processing, multi-slice 2D MRI still accounts for the majority of clinical scans because it: 1)~is faster to acquire; 2)~offers clinically relevant tissue contrasts difficult to acquire in 3D~\cite{bazot2013comparison}; 3)~has relaxed hardware requirements; 4)~is very familiar to clinicians; and 5)~can be less sensitive to motion artifacts.
For these reasons, multi-slice 2D MRI are common in pragmatic clinical trials~\cite{treatms_trial} and in pre-clinical imaging studies~\cite{stringer2021review}.

Multi-slice 2D MR neuroimages can be strikingly anisotropic.
Although in-plane resolutions for 2D MRI are frequently less than 1~mm in both phase-encode and readout directions, slice thicknesses may be 3--8~mm and many acquisitions additionally have slice gaps of 0.5--2~mm.  
Automated processing pipelines, which are often designed and trained for isotropic image volumes, perform sub-optimally on such anisotropic volumes~(even when interpolated to isotropic voxel spacing). 
Super-resolution~(SR) aims to solve this problem by estimating high-resolution~(HR) images from low-resolution~(LR) acquisitions. 
Most contemporary work on SR for MRI focuses on deep neural networks.
Convolutional neural networks~(CNNs) form the backbone of many approaches~\cite{sood_anisoSRGAN_2019,du2020super}, including generative adversarial networks~(GANs)~\cite{zhang2022soup,chen2020mri,lyu2020mri}, normalizing flows~\cite{lugmayr2020srflow,kim2021noise,liu2023simultaneous,ko2023mriflow}, and denoising diffusion probabilistic models~(DDPMs)~\cite{wu2023super,wang2023inversesr,wu2024anires2d,chang2024high,han2024arbitrary,wu20243d}. 
The transformer architecture~\cite{lu2022transformer,yang2020learning,yan2021smir,li2022transformer,feng2021task} is also common, sometimes used alongside CNNs~\cite{zhou2023hybrid,forigua2022superformer,cheng2023hybrid,fang2022hybrid}.

\begin{figure*}[!t]
    \centering
    \includegraphics[width=0.9\linewidth]{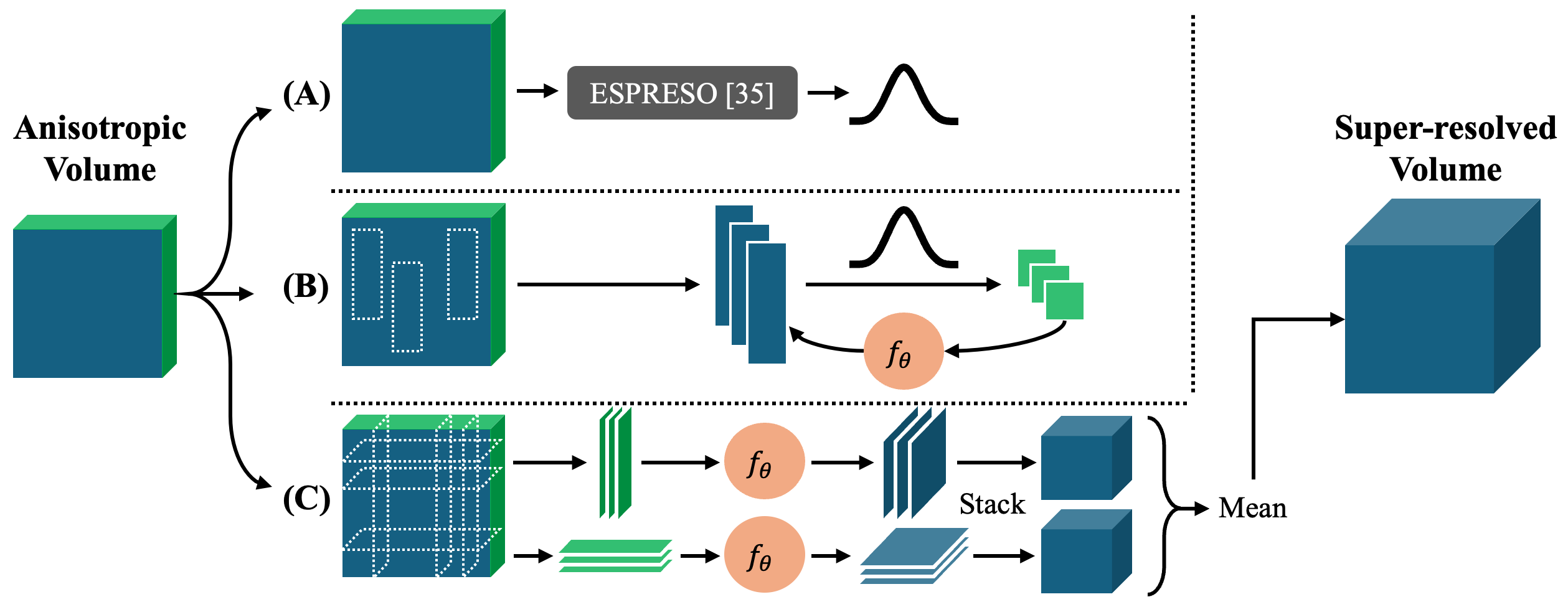}
    \caption{A flowchart of our proposed method. The anisotropic input volume is fed independently into each of the three steps. First, in (A)~(Sec.~\ref{sec:approx_imaging}), we estimate the slice excitation profile with ESPRESO~\cite{han2023espreso}. Second, in (B)~(Sec.~\ref{sec:eclare_training}), we extract HR in-plane 2D patches and use the PSF estimated from (A) to create paired training data. This training data is used to train the network $f_\theta$ with supervised learning. Third, in (C)~(Sec.~\ref{sec:apply_model}), we extract LR through-plane 2D slices and super-resolve them with the trained network $f_\theta$ from (B).The super-resolved slices are stacked and averaged, yielding the super-resolved output volume.}
    \label{fig:method}
\end{figure*}

Although many SR approaches demonstrate remarkable performance, none at present satisfy all desirable properties for SR of multi-slice 2D MRI.
Most methods, for example, do not consider the slice profile and true through-plane resolution of 2D MRI.  
Indeed, many approaches conflate slice separation with through-plane resolution, but they are different concepts and must be handled differently. 
For example, methods evaluated on simulations that use LR data that is only downsampled from 3D volumes~(i.e., with interpolation) and without proper simulation of slice profile and slice separation would be more properly labeled as ``inpainting'' or ``slice imputation'' rather than super-resolution. 
As well, many methods in the literature have been trained for super-resolution of single integer factors such as 2 and 4~\cite{sood_anisoSRGAN_2019,zhang2018image,lugmayr2020srflow,wu2023super,lu2022transformer,zhou2023hybrid}. 
Not only do these methods require retraining for each integer factor that is encountered, they also use techniques like power-of-two upsampling or pixel shuffle~\cite{shi2016pixelshuffle} that do not allow for non-integer factors. 
Finally, many methods are trained on specific MR tissue contrasts and often~(in neuroimaging) on skull-stripped data, necessitating retraining and possibly re-evaluation for alternative tissue contrasts and for with-skull data.
\textit{At present, there is no algorithm that handles all of these concerns for an arbitrary multi-slice 2D MR image volume.}

To address these concerns, we propose ECLARE~(Efficient Cross-planar Learning for Anisotropic Resolution Enhancement), a deep learning algorithm to super-resolve multi-slice 2D MR volumes along the through-plane direction.
ECLARE builds on a succession of self-SR approaches~\cite{jog2016self,zhao2020smore,remedios2021joint,remedios2023self}, in which all training data is extracted from the HR in-plane slices within a LR image volume.
However, ECLARE adds several crucial features to properly model slice acquisition, account for non-integer super-resolution factors, and incorporate efficient network design elements. 
In particular, we use ESPRESO~\cite{han2023espreso}, a GAN-based algorithm previously developed by our team, to estimate the relative slice selection profile.
We use this profile to model slice thickness and gap and to simulate training data from HR in-plane patches extracted from the anisotropic input volume.
These data are used to train an efficient SR neural network architecture based on ``wide activation SR''~(WDSR)~\cite{yu2018wide}.
The model is then applied to both canonical through-planes, yielding two super-resolved volumes that are then averaged to create the final output. 
Underlying all resampling operations in our approach is a novel field-of-view~(FOV)-aware resampling algorithm that respects the image coordinate system and target sample spacing.
The main contributions of this paper are the following:

\begin{itemize}
    \item We propose an FOV-aware resampling to regrid an image while preserving a desired sample spacing and maintaining the location of the FOV.
    \item We model the relationship between isotropic and anisotropic volumes with explicit slice thickness and slice gap, then leverage ESPRESO to learn this model.
    \item We propose an efficient alternative to previous self-SR approaches in MR: a single-model architecture without pre-trained weights to perform both anti-aliasing and SR.
    \item We propose an architectural change to WDSR to yield the correct number of slices at the correct FOV at both integer and non-integer anisotropy factors.
    \item We evaluate our approach and compare against several methods in both signal recovery metrics and downstream task performance in two disparate datasets.
\end{itemize}
Our proposed method super-resolves an anisotropic volume without external training data and can serve as an alternative to spatial interpolation with improved quality.

\section{Related Work}
In this work, we consider SR for 2D multi-slice MRI.
While there is a body of literature discussing SR approaches focusing on training a model with external data, we do not discuss it here.
Such approaches create paired LR-HR data through simulation from large datasets, perform supervised learning with or without consideration of arbitrary scale factors, and either must be applied to a similar domain as the training set or undergo domain adaptation for practical use.
Alternatively, in this paper we consider methods that aim to be applicable ``out-of-the-box.'' 
Accordingly, our discussion of related work is limited to self-supervised or foundational models for SR of 2D anisotropic MR volumes.
Thus, at test-time, the existence of relevant external training data is not assumed.

\begin{figure*}[!t]
    \centering
    \includegraphics[width=0.9\linewidth]{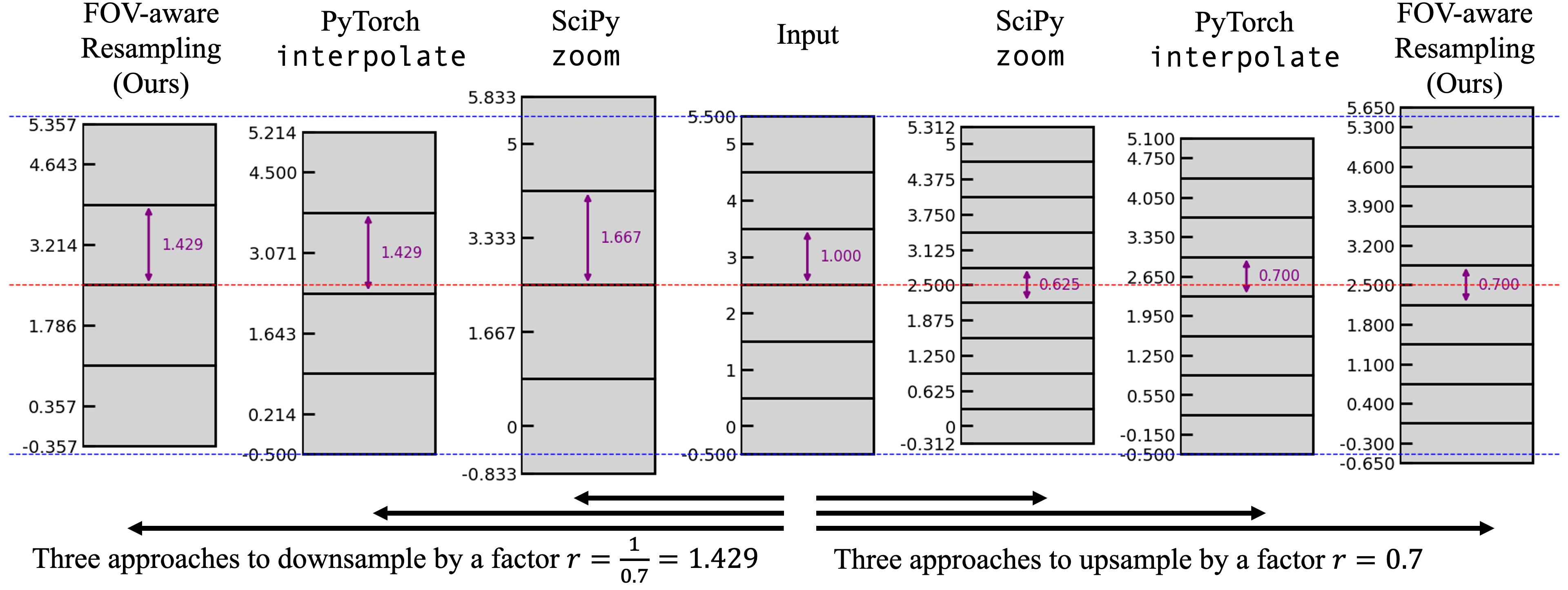}
    \caption{A 1D demonstration of the differences between resampling implementations. The input has 6 pixels and its FOV center is located at 2.5, indicated by the red dashed line. Its sample spacing is 1, indicated by the purple arrow. Its extents occur at -0.5 and 5.5, indicated by the blue dashed line. In this demonstration, the desired sample spacing for upsampling is 0.7, and the desired sample spacing for downsampling is 1.429. The implementation \texttt{scipy.ndimage.zoom}~\cite{scipy_zoom} maintains the FOV center and boundary sample locations, but changes the sample spacing to 0.625 and 1.667 for upsampling and downsampling, respectively. The implementation \texttt{torch.nn.functional.interpolate}~\cite{pytorch} anchors the FOV extent to start at 0.5, shifting the center from the desired FOV center. Our proposed method~(Sec.~\ref{sec:resampling}) maintains both the sample spacing and the FOV center.}
    \label{fig:resampling}
\end{figure*}

Self-SR in MRI was first explored~\cite{jog2016self} by fitting an anchored neighborhood regression~\cite{timofte2013anchored} model to learn the mapping from LR to HR patches, followed by Fourier burst accumulation~(FBA)~\cite{delbracio2015removing} to aggregate estimated high frequency Fourier coefficients into the super-resolved volume. 
The related work SMORE~\cite{zhao2020smore} reinterpreted self-SR using CNNs, which greatly increased the output accuracy and visual appearance. 
SMORE trains two enhanced deep residual networks~(EDSRs)~\cite{lim2017enhanced}: one for anti-aliasing and one for SR.
These networks are applied in cascade to orthogonal through-plane slices, and all intermediate results are fused with FBA.
Furthermore, although SMORE can be a self-SR method, \cite{zhao2020smore} found that pre-training on similar data~(i.e., human head MR images) and fine-tuning on the anisotropic volume at test-time allowed fewer training iterations and achieved better results.

Implicit neural representations~(INR)~\cite{mildenhall2020nerf,sitzmann2020implicit,wu2022arbitrary} learn the mapping from coordinate to intensity.
Since coordinates can be sampled at any resolution, this has inspired several works related to arbitrary-scale SR~\cite{tan2020arbitrary,wu2022arbitrary,liu2024arbitrary}.
Recently, multi-contrast INR~\cite{mcginnisandshit2023inr} was proposed to super-resolve subjects with multiple available contrasts from the same session, ideally acquired at different orientations (e.g., sagittal \TOne-w and coronal \TTwo-w). 
However, that method also supports the use of a single contrast, training in a similar manner to self-SR methods.

SynthSR~\cite{iglesias2023synthsr} is a joint synthesis and SR model.
Training data for SynthSR includes real MPRAGE head volumes and their corresponding parcellations.
Domain randomization uses a Gaussian mixture model to produce synthetic MR-like images of various contrasts alongside augmentations such as simulated anisotropy, slice separations, bias fields, and additive noise.
Through domain randomization, SynthSR maps an MR image of arbitrary contrast to a 1~mm isotropic \TOne-w MPRAGE scan.

\section{Methods}
\label{sec:methods}
Our proposed ECLARE improves resolution along the through-plane axis of an anisotropic 2D-acquired multi-slice MR image volume.
To explain ECLARE, we first model the relationship between HR and LR volumes~(Sec.~\ref{sec:modeling_imaging}).
We then describe how to approximate the slice selection profile~(Sec.~\ref{sec:approx_imaging}) and FOV-aware resampling~(Sec.~\ref{sec:resampling}).
We approximate the slice selection profile with ESPRESO~\cite{han2023espreso} and use it to train a CNN with in-plane HR patches and their simulated LR counterparts~(Sec.~\ref{sec:eclare_training}).
Finally, we show how ECLARE uses this 2D CNN to super-resolve the 3D volume~(Sec.~\ref{sec:apply_model}) and describe implementation details~(Sec.~\ref{sec:implementation}).

\begin{figure*}[!t]
    \centering
    \includegraphics[width=0.9\linewidth]{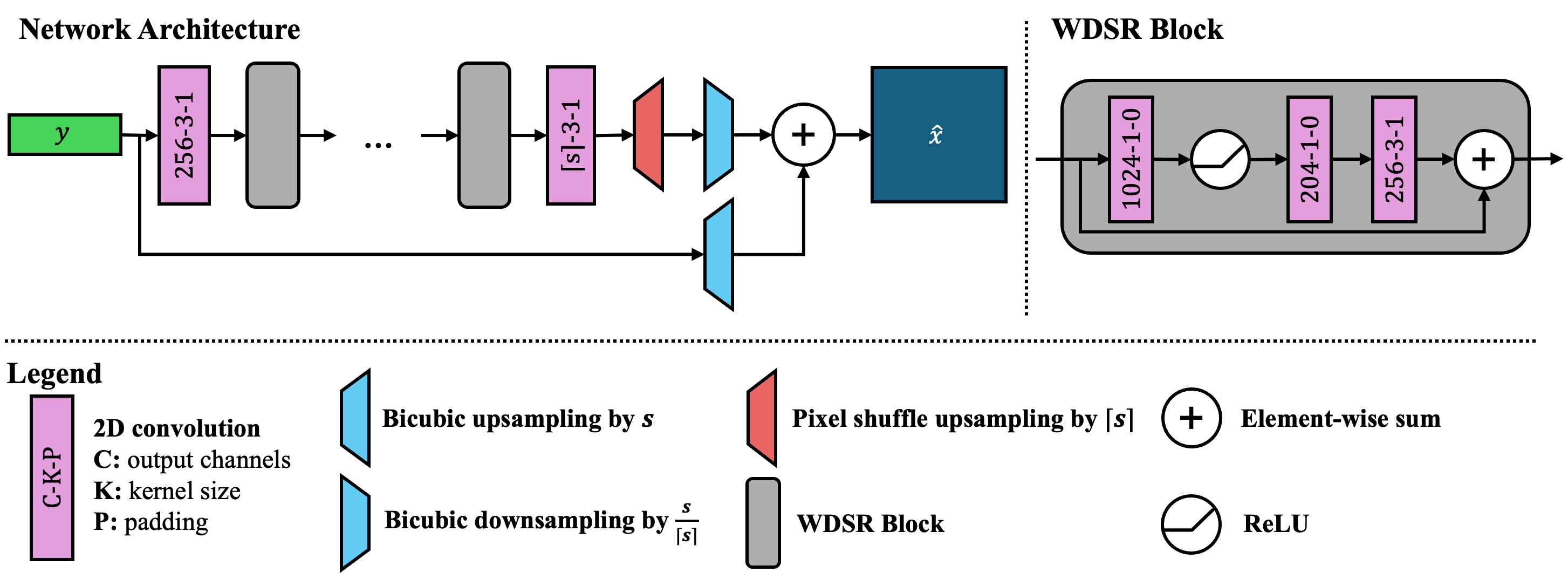}
    \caption{A diagram of our proposed architecture, based on WDSR~\cite{yu2018wide}. The low resolution image $y$ is first processed by a convolution to produce a feature map with 256 channels. It then passes through several WDSR blocks. A final convolution is done to produce the correct number of channels for pixel shuffle. Finally, the result is downsampled and added to an upsampled $y$ to yield the output. In our implementation, we use 16 WDSR blocks.}
    \label{fig:architecture}
\end{figure*}

\subsection{Modeling the imaging process}
\label{sec:modeling_imaging}
A 2D-acquired MR image volume is a stack of independent 2D slices.
In-plane, data is acquired in k-space and transformed to the image domain through the inverse Fourier transform.
For through-plane, the relationship between slices is entirely characterized in the image domain.
Common slice profiles take the shape of truncated Gaussians or use the Shinnar-Le Roux algorithm~\cite{pauly1991parameter,shinnar1989synthesis} to approximate a rect.
All in-plane measurements are subject to the same slice thickness and location.

We denote the slice selection profile as $h$.
It is a continuous function of the through-plane position.
We note that although slice selection is implemented as a bandpass filter, relative to a particular through-plane axis position the slice profile is a low-pass filter.
This corresponds to the observation that ideal slice profiles are rect-like or Gaussian-like.

An ideal multi-slice 2D MR volume $x$ would be acquired in an isotropic fashion with resolution $a\times a \times a$~mm.
For practical reasons~(time and physical constraints), however, a real multi-slice 2D MR volume $y$ is acquired with thicker slices---i.e., $a \times a \times b$~mm where $b > a$.
A further problem is that $y$ may have slice gaps where the slice separation in the through-plane direction is greater than $b$, the slice thickness.
We model 2D multi-slice acquisitions as a discrete convolution between the unobserved HR image volume and the slice selection profile:
\begin{equation}
    y = (x \ast h)\downarrow_r,
\label{eq:forward}
\end{equation}
where $\downarrow_r$ is the subsampling ratio of $r$.
The full-width-at-half-max~(FWHM) of $h$ determines the slice thickness and $r$ is a ratio between the HR slice separation~$s^\ast$ and the LR slice separation~$s$, defined as $r = \frac{s}{s^\ast}$.
If $h$ has a smaller FWHM than the separation $s$, then Eq.~\ref{eq:forward} models slice gap.
Similarly, a FWHM of $h$ equal to or greater than $s$ respectively models adjacent slices and slice overlap.

\subsection{Approximating the imaging process}
\label{sec:approx_imaging}
The imaging process described by Eq.~\ref{eq:forward} comprises blurring with $h$ and subsampling by $r$.
The subsampling factor $r$ is known since the HR and LR slice separations are each known, so the remaining task is to determine the slice selection profile $h$.
In general, $h$ is a product of a real-world implementation that is subject to imperfections in the instrumentation, and hence cannot be derived precisely.
However, since our intent is to use HR in-plane patches as the target distribution, estimating $h$ directly is not necessary.
Instead, we leverage ESPRESO~\cite{han2023espreso} to estimate a \textit{relative} slice selection profile $\tilde{h}$, shown in Fig.~\ref{fig:method}~(A).
The desired relative slice selection profile $\tilde{h}$ blurs data from the in-plane slices such that the distribution matches the LR through-plane data.
To estimate $\tilde{h}$, ESPRESO optimizes a neural network to produce a fixed-size discrete profile through adversarial training. 
It trains on a single LR anisotropic volume by assuming that degraded in-plane and through-plane patches are indistinguishable in distribution for an appropriate patch size.
Briefly, training consists of extracting in-plane image patches, degrading by $\tilde{h}$, comparing to through-plane patches, and backpropagating the loss through a discriminator. 
The optimal $\tilde{h}$ yields degraded in-plane image patches that are indistinguishable in distribution from true through-plane image patches.
We use this estimated $\tilde{h}$ in lieu of $h$ for Eq.~\ref{eq:forward} hereafter.

\subsection{FOV-aware resampling}
\label{sec:resampling}
While resampling is routine in digital signal processing, the exact implementation has several freedoms.
ECLARE requires both downsampling~(for training data simulation) and upsampling~(for inference) and we would like both to be stable with respect to the FOV.
We consider three conditions that are desirable to meet with respect to the FOV of the signal:
\begin{enumerate}
    \item the center of the FOV should be at the same location;
    \item the samples should be spaced as specified;
    \item the locations of boundary samples should be maintained.
\end{enumerate}
Resampling algorithms cannot meet all of these conditions in general since only a discrete number of samples is permitted.
However, since it is possible to satisfy two of these conditions, a compromise on the third is necessary.
We quickly review popular modern implementations.
The method \texttt{scipy.ndimage.zoom}~\cite{scipy_zoom} meets conditions 1) and 3). It adjusts the sample spacing to preserve the FOV center while locking the locations of boundary samples.
The method \texttt{torch.nn.functional.interpolate}~\cite{pytorch} meets only condition 2). It preserves the sample spacing, but neither the FOV center nor the boundary sample locations.
Instead, it anchors the extent on the first sample.
We place more importance on the center of the FOV and the sample spacing and therefore desire conditions 1) and 2).
We summarize the resampling method proposed in~\cite{han2022thesis} that maintains the FOV center and sample spacing while sacrificing the precise location of the boundary samples.
A comparison of the three implementations is shown in Fig.~\ref{fig:resampling}.

\begin{table}[h]
    \centering
    \rowcolors{2}{gray!20}{white} 
    \caption{
    Quantitative results for the ablation study. For brevity, E denotes ESPRESO, F denotes FOV-aware resampling, and W denotes the modified WDSR architecture. The model absent of any contributions is Baseline, and the model using all contributions is ECLARE. See Sec.~4.2 for complete details. PSNR and SSIM are reported as mean $\pm$ standard deviation. Best results are \textbf{bolded} and second-best results are \underline{underlined}.
    }
    \begin{tabular}{cccccc}
    \toprule
        Configuration & E      & F      & W      & PSNR                       & SSIM\\
    \midrule
        Baseline      & \xmark & \xmark & \xmark & $27.72\pm0.84$             & $0.84\pm0.02$\\
                      & \xmark & \cmark & \xmark & $28.00\pm0.86$             & $0.85\pm0.02$\\
                      & \xmark & \xmark & \cmark & $27.91\pm0.86$             & $0.85\pm0.02$\\
                      & \cmark & \xmark & \xmark & $27.48\pm2.80$             & $0.80\pm0.16$\\
                      & \cmark & \xmark & \cmark & $29.28\pm0.85$             & \underline{$0.87\pm0.02$}\\
                      & \cmark & \cmark & \xmark & \underline{$29.32\pm0.85$} & $0.86\pm0.02$\\
                      & \xmark & \cmark & \cmark & $28.61\pm0.87$             & \underline{$0.87\pm0.02$}\\
        ECLARE        & \cmark & \cmark & \cmark & $\mathbf{30.29\pm0.87}$ & $\mathbf{0.88\pm0.02}$\\
    \bottomrule
    \end{tabular}
    
    \label{tab:ablation}
\end{table}

\begin{figure}[!t]
    \centering
    \includegraphics[width=\linewidth]{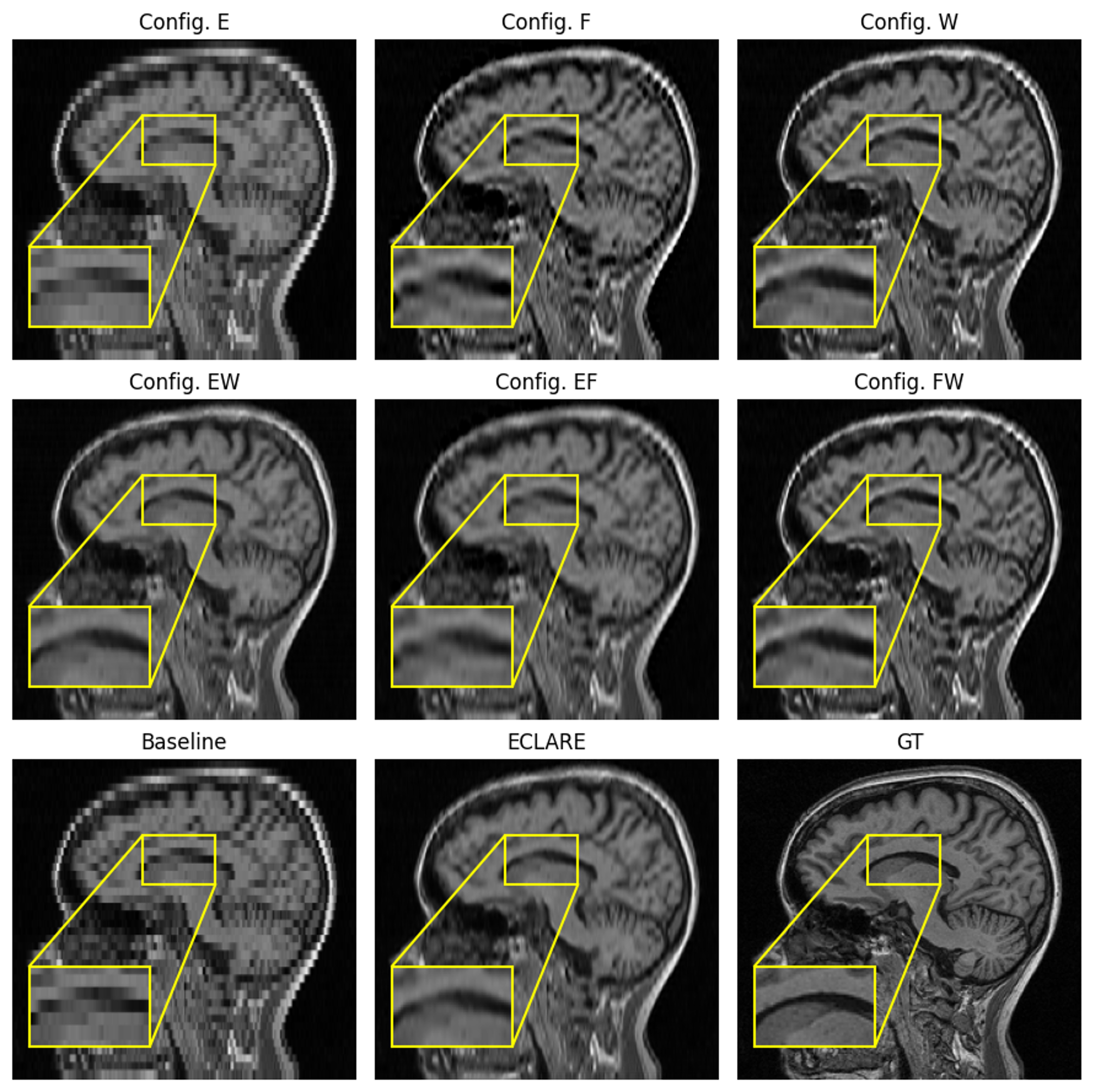}
    \caption{
    Ablation study results labeled by configuration. E denotes ESPRESO, F denotes FOV-aware resampling, and W denotes the modified WDSR architecture. A sagittal slice of a representative subject is shown, and a zoomed inset indicates regions where the absence of ESPRESO in the algorithm yields a hyperintense artifact.
    }
    \label{fig:ablation}
\end{figure}

\begin{figure*}[!t]
    \centering
    \includegraphics[width=\linewidth]{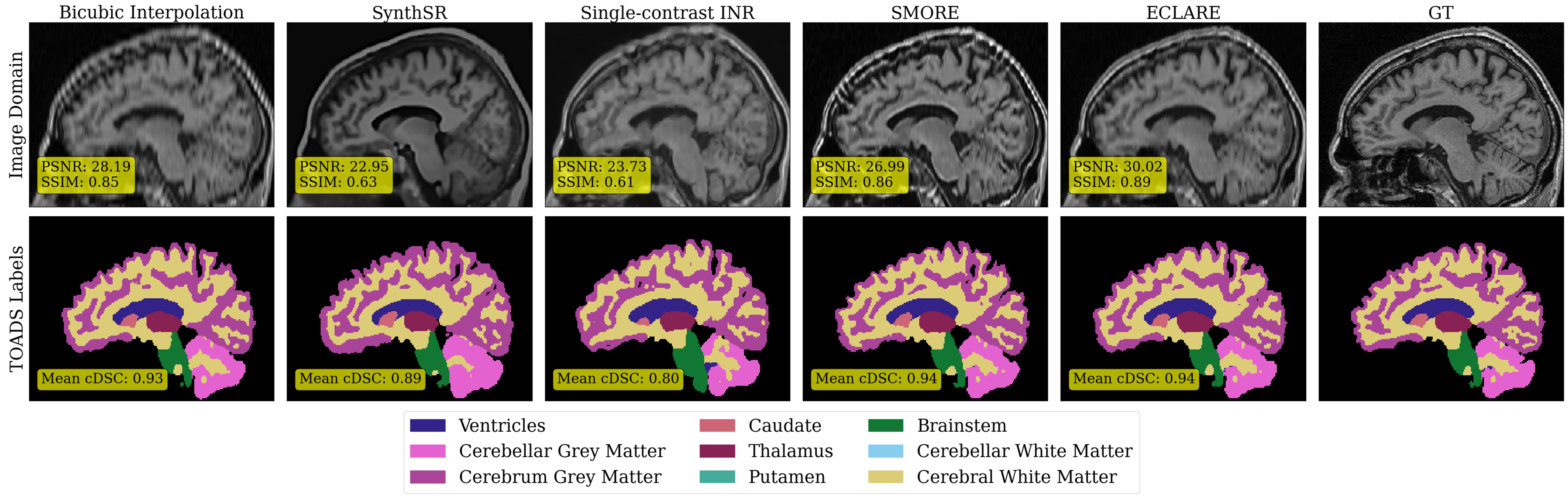}
    \caption{A $5\|1.5$ sagittal slice from a representative subject from the OASIS-3 dataset. The first row shows the image domain with PSNR and SSIM for the entire image volume overlaid. The second row shows the simplified SLANT labels, with an overlay showing the mean cDSC across all labels for the entire image volume.}
    \label{fig:qual-oasis}
\end{figure*}

As in~\cite{han2022thesis}, we introduce a fixed coordinate system for the signal before and after resampling.
Let the position of the first pixel be centered at zero, let $N$ be the number of pixels before resampling, and let the spacing before resampling be one. 
The width of each pixel is equal to the spacing; before resampling it is therefore also one.
The first pixel $p_0$ is centered at zero and the last pixel $p_{N - 1}$ is centered at $N - 1$. 
Since each pixel has a width of one, $p_0$ extends from $-0.5$ to $0.5$ and $p_{N - 1}$ extends from $N - 1.5$ to $N - 0.5$.
Therefore the entire image extends from $-0.5$ to $N - 0.5$.

We now calculate the new positions and extents of pixels after resampling with respect to the same coordinate system.
Let the new spacing~(and pixel width) be $d$.
After resampling, the new image must have an integer number of pixels.
With $N^\prime$ as the number of pixels after resampling and $\lfloor \cdot \rceil$ as the rounding operator,
\begin{equation}
    N^\prime = \left\lfloor \frac{N}{d} \right\rceil.
\end{equation}
Accordingly, we notate the first pixel after resampling as $p^\prime_0$ and the last pixel as $p^\prime_{N^\prime-1}$.

In order to center around the same FOV, the center position of the entire image should be identical before and after resampling. 
Since the center of the FOV is the halfway point between the first and last pixels, we can write
\begin{align}
    c = \frac{1}{2}(p_0 + p_{N - 1}) &= \frac{1}{2}(p^\prime_0 + p^\prime_{N^\prime - 1})
\end{align}
Also, since we know the new spacing $d$ and the new number of pixels $N^\prime$, we can calculate the position of any pixel $p_i^\prime$, $i=0, 1, 2,\ldots,N^\prime-1$ in terms of the new starting pixel:
\begin{equation}
    p^\prime_i = p^\prime_0 + di.
\end{equation}
To find the position of the first pixel $p^\prime_0$, we use the center $c$:
\begin{align}
    c &= \frac{1}{2}(p^\prime_0 + p^\prime_{N-1})\\
      &= \frac{1}{2}(p^\prime_0 + [p^\prime_0 + d(N^\prime - 1)])\\
    p^\prime_0 &= -\frac{d}{2}(N^\prime - 1) + c.
\end{align}
Finally, we can calculate the new extents of the entire image after resampling. The first pixel extends from $p^\prime_0 - \frac{d}{2}$ to $p^\prime_0 + \frac{d}{2}$ and the last pixel extends from $p^\prime_{N-1} - \frac{d}{2}$ to $p^\prime_{N-1} + \frac{d}{2}$. Thus, the image extents are $p^\prime_0 - \frac{d}{2}$ to $p^\prime_{N-1} + \frac{d}{2}$.

The result of this method yields an FOV-aware resampling implementation that maintains the FOV center and guarantees the desired pixel spacing.
This is true in both upsampling and downsampling scenarios as demonstrated in Fig.~\ref{fig:resampling}.

\subsection{Training a deep learning model with paired data}
\label{sec:eclare_training}
We now develop a relationship between in-plane and through-plane data and train a SR network.
Paired LR and HR data are necessary for supervised learning.
We extract HR 2D patches from in-plane and produce blurred 2D patches using Eq.~\ref{eq:forward} with the learned $\tilde{h}$ from ESPRESO.
These blurred 2D patches are then downsampled using the FOV-aware resampling described in Sec.~\ref{sec:resampling}.
Thus, each LR patch has the appropriate simulated slice thickness, separation, and FOV.

A deep convolutional neural network model $f_\theta$ learns the mapping between the paired patches, as shown in Fig.~\ref{fig:method}~(B).
We use an architecture based on WDSR~\cite{yu2018wide} after empirically evaluating other architectures such as EDSR~\cite{lim2017enhanced} and RCAN~\cite{zhang2018image}.
The WDSR architecture, designed for efficiency, achieves the same quantitative metrics as other architectures but trains in less time.
However, the WDSR architecture itself uses pixel shuffle~\cite{shi2016pixelshuffle}, which requires integer upsampling factors, to perform the final upsampling to the desired image size.
In general, though, the required upsampling ratio $r$ may be non-integer. 
To address this, we rewrite $r$ as
\begin{equation}
    r = \lceil r \rceil \frac{r}{\lceil r \rceil},
\end{equation}
where $\lceil\cdot\rceil$ is the ceiling operation.
Accordingly, to achieve the desired $r$, we first upsample by $\lceil r \rceil$ and then downsample by $\frac{r}{\lceil r \rceil}$.
Upsampling is done with pixel shuffle and downsampling is done with FOV-aware resampling~(Sec.~\ref{sec:resampling}).
Our model is illustrated in Fig.~\ref{fig:architecture}.

\subsection{Inference}
\label{sec:apply_model}
The trained network is fully convolutional, allowing it to be used on full-sized image slices despite training on patches.
Our inference strategy to super-resolve the volume is to independently super-resolve all slices from each canonical through-plane axis, stack the super-resolved slices into two volumes, and take an unweighted mean of the two volumes to produce the final output.
This procedure is illustrated in Fig.~\ref{fig:method}~(C).

\begin{figure*}[!t]
    \centering
    \includegraphics[width=\linewidth]{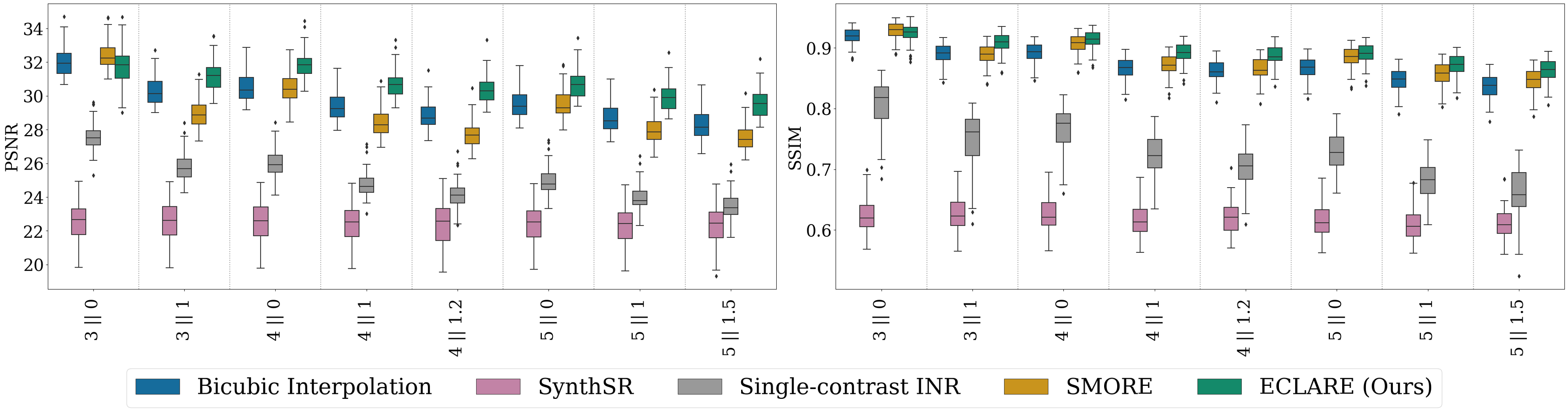}
    \caption{Quantitative results~(PSNR on the left, SSIM on the right) on the OASIS-3 dataset across resolutions.}
    \label{fig:psnr-ssim-oasis}
\end{figure*}

\subsection{Implementation details}
\label{sec:implementation}
ECLARE is implemented in PyTorch and we train the model from scratch with each new multi-slice 2D MR image.
We optimize the parameters of the network by minimizing mean squared error using a total of $1,000,000$ patches with a batch size of $128$ patches.
The training patches are of size $\lceil8r\rfloor\times8$.
We use a $21$-tap filter to represent $h$, which follows the standard implementation of ESPRESO.
We used the Adam~\cite{kingma2014adam} optimizer, a learning rate of $10^{-3}$, the one-cycle learning rate scheduler~\cite{smith2019onecycle}, and PyTorch's automatic mixed precision feature. 
Our implementation uses $16$ residual blocks of $256$ channels.
With these hyperparameter choices, on an NVIDIA V100 the total running time, including self-training and inference, is just under 5 minutes.
The memory usage is 1.16 GB on the GPU.
Our code is open-source and is available at \href{https://www.github.com/sremedios/eclare}{https://www.github.com/sremedios/eclare}.

\section{Experiments and Results}
\label{sec:experiments_and_results}

\subsection{Data and evaluation metrics}
\label{sec:data_and_eval_metrics}
Experiments using pairs of 2D-acquired LR images and 3D-acquired HR images would be ideal.
However, quantitative analysis is not feasible without simulations.
Acquiring high-resolution 2D scans increases scan time linearly unless methods to reduce scan time are used.
Longer scan times and acceleration techniques can introduce artifacts that would be detrimental to the analysis.
3D scans are quicker, but acquire data differently and sometimes require different contrasts, eliminating possible comparisons.
This motivates a physics-based simulation approach, established as a precedent previously~\cite{pauly1991parameter,shinnar1989synthesis,jog2016self,zhao2018deep,zhao2019applications,zhao2020smore,han2021espreso,han2023espreso}.

Our experiments began with HR isotropic 3D-acquired image volumes of resolution $1\times1\times1\text{~mm}^3$. 
These images were then degraded with Eq.~\ref{eq:forward}, setting $h$ to the shape of the slice excitation profile created by the Shinnar-Le Roux~(SLR) pulse design algorithm~\cite{pauly1991parameter,shinnar1989synthesis}, $s$ to the desired slice spacing~(described below), and $x$ to a 3D-acquired isotropic volume. 
The implementation of the SLR algorithm we use is from the \texttt{sigpy} library~\cite{martin2020sigpy}. 

The physical dimensions of the slice thickness is determined by the FWHM of the slice selection profile.
Since our images began with $1\text{~mm}$ thick slices, this involved choosing the FWHM of $h$ to be equal to the desired slice thickness.
In clinical neuroimaging, it is common to acquire no slice gap or a slice gap equal to $1\text{~mm}$ or $1/3$ the slice thickness, so we chose to simulate 2D acquisitions of thicknesses $3, 4,\text{ and }5\text{~mm}$ with gaps $0, 1, 1.2,\text{ and }1.5\text{~mm}$, yielding a total of $8$ resolutions. 
We adopt the notation $T\|G$~\cite{remedios2023self} for a thickness of $T$~mm and gap of $G$~mm and list our choices here: $(3\|0), (3\|1), (4\|0), (4\|1), (4\|1.2), (5\|0), (5\|1),\text{ and }(5\|1.5)$. 
These simulation experiments were conducted on two datasets. The first consisted of $50$ \TOne-w images from healthy subjects in OASIS-3~\cite{LaMontagne2019_OASIS3}. 
The second consisted of $30$ \TTwo-w FLuid Attenuated Inversion Recovery~(FLAIR) images from people with MS in the 3D-MR-MS dataset~\cite{lesjak2018novel}.

We compare ECLARE to bicubic interpolation, SynthSR~\cite{iglesias2023synthsr}, single-contrast INR~\cite{mcginnisandshit2023inr}, and SMORE~\cite{zhao2020smore}. 
Since its output is always a \TOne-w MPRAGE, we do not evaluate SynthSR on the FLAIR SR task.
Other mainstream SR methods are not appropriate for comparison here due to the heterogeneities in data acquisition strategies for multi-slice 2D MRI.
Many LR volumes have non-integer ratios to achieve isotropic resolution or have slice gap.
Furthermore, there is a need to evaluate on arbitrary image contrasts without retraining for each subtle change in contrast.
ECLARE and the methods we compare to are agnostic to these differences.

Since we have ground truth~(GT) images for simulated data, we can use reference-based image quality metrics to evaluate SR performance. 
We use peak signal-to-noise ratio~(PSNR) and structural similarity index measure~(SSIM) with the implementations provided by the \texttt{scikit-image} Python library~\cite{van2014scikit}.
We additionally evaluate the utility of the super-resolved images by segmentation downstream tasks. 
We use the \textit{consistency Dice similarity coefficient}~(cDSC), which we define as the Dice similarity coefficient~(DSC) between an automated algorithm's performance on the HR image volume and the super-resolved image volume. 
This is to differentiate cDSC from DSC which commonly is used between an algorithm-generated label and a human-generated label.

\begin{figure*}[!t]
    \centering
    \includegraphics[width=\linewidth]{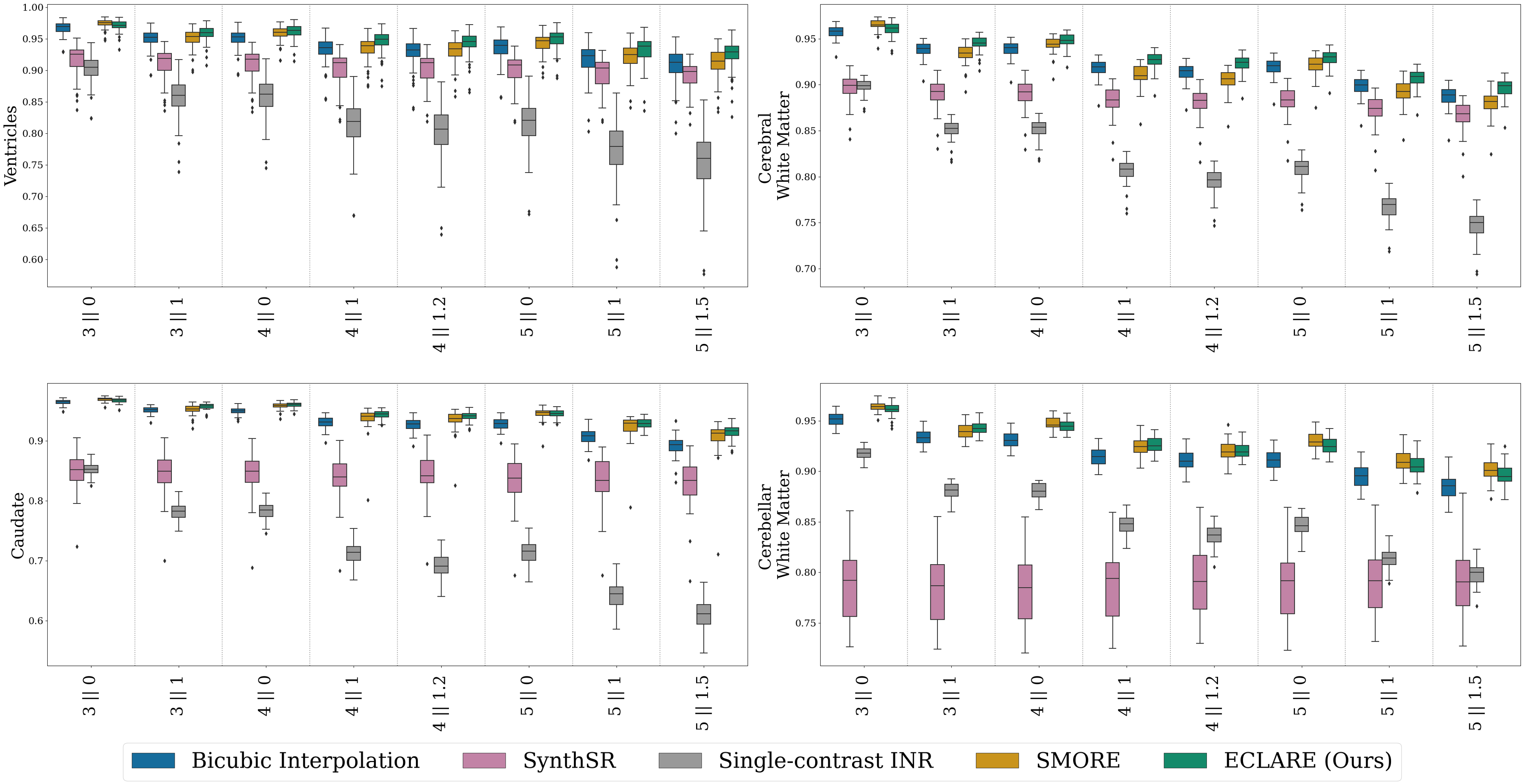}
    \caption{Quantitative results on cDSC by anatomical region on the OASIS3 dataset across resolutions. In the interest of space, only four representative regions (namely, ventricles, cerebral white matter, caudate, and cerebellar white matter) are shown.}
    \label{fig:cdsc-oasis}
\end{figure*}

\subsection{Ablation on components}
\label{sec:ablation}

In this section, we perform an ablation to validate components of our method.
The ablation cohort consisted of $50$ subjects from the OASIS-3 dataset at the resolution $4\|1.2$.
Since we have three primary technical contributions~(FOV-aware resampling, the use of ESPRESO to estimate slice selection profiles, and modifications to the WDSR architecture for non-integer scale factors), there are eight configurations to examine.
When FOV-aware resampling is absent, \texttt{torch.interpolate} with cubic B-splines is used instead.
When ESPRESO is absent, a Gaussian slice selection profile with thickness equal to the slice separation is used instead.
When the WDSR architectural modifications are absent, resampling is used instead of pixel shuffle.
(This choice of resampling is determined by the corresponding ablation and refers to either FOV-aware resampling or \texttt{torch.interpolate}.)
The Baseline configuration omits all of these changes, while the ECLARE uses all contributions.

We evaluated each configuration with PSNR and SSIM, shown in Table~\ref{tab:ablation}.
By both metrics, ECLARE outperformed all other configurations.
With the Wilcoxon signed-rank test and Bonferroni correction for multiple comparisons, the improved performance of ECLARE over all other ablations was significant~($p \ll 0.001$).
However, the ablation study revealed some interesting contributions.
First, we will discuss the single-contribution configurations.
In all three cases, results did not substantially improve from the baseline for PSNR and SSIM.
Interestingly, using ESPRESO alone without FOV-aware resampling or the WDSR pixel shuffle modifications reduced quantitative performance; note how the PSNR score exhibits not only lower average PSNR but higher PSNR standard deviation as well.
This potentially could signify the sensitivity of ESPRESO to the resampling procedure, which is indicated by the two-contribution ablations.
For the two-contribution ablations, performance improved in all cases compared to the Baseline.
The combination of ESPRESO with either of the other two contributions yielded the highest PSNR results, and the scores for SSIM are comparable across all three two-contribution ablations.

Qualitative results of a representative subject for the ablation study are shown in Fig.~\ref{fig:ablation}.
Both the baseline and using ESPRESO alone failed to recover from the blocky aliasing artifacts.
The inclusion of ESPRESO is important, however.
Note the regions indicated by the zoomed insets; when ESPRESO is absent from a configuration, hyperintense artifacts appear at the interfaces between dark and light intensities.
However, when ESPRESO is present, those hyperintense artifacts are not apparent.


\subsection{Healthy \texorpdfstring{\TOne}{T1}-w simulations}
\label{sec:oasis}

We calculated cDSC on the SLANT~\cite{huo20193d} and SynthSeg~\cite{billot2023synthseg} estimates of the OASIS-3 dataset. Since SLANT produces many small labels, we simplify them down to $9$ labels as in our previous work~\cite{remedios2023self}: ventricles, cerebellar gray matter~(GM), cerebellar white matter~(WM), cerebral GM, cerebral WM, caudate, putamen, thalamus, and brainstem. SynthSeg is a generalized automated segmentation technique designed to be robust to a variety of image appearances. It directly produces volume estimates of similar regions of interest, so we include the following comparable regions: ventricles, cerebral white matter, cerebral gray matter, cerebellar white matter, cerebellar grey matter, thalamus, putamen, caudate, and brainstem.

On the OASIS-3 dataset, we show qualitative results from a representative subject at $5\|1.5$ resolution in Fig.~\ref{fig:qual-oasis}. 
The image domain is shown alongside the SLANT result labels across the methods.
In the image domain, it is apparent that our proposed ECLARE method is visually closer to the ground truth~(GT) in both appearance and contrast.
Viewing Fig.~\ref{fig:qual-oasis} column-by-column, in the image domain bicubic interpolation does not address the extreme aliasing, but is relatively accurate with respect to the SLANT labels. 
SynthSR generates a realistic brain and head appearance, but mismatches the contrast and hallucinates some structures such as the cortical sulci that can be seen in the SLANT labels row. 
Single-contrast INR does not generate accurate results, synthesizing a different brain stem appearance from the other methods, as shown in the green brainstem label in the second row. 
The SMORE results highlight an ``over-sharpening'' effect with hyperintense regions near the skull and superior to the ventricles and the bottom part of the brainstem. 
ECLARE yields the most visually accurate image of the group in both the image as well as qualitative inspection of the segmentation.
The SLANT segmentation results are still not ideal for any method, and some areas of cortical gray matter are missed by all methods.

We show the PSNR and SSIM for the five methods in Fig.~\ref{fig:psnr-ssim-oasis}. 
Across all resolutions, our proposed ECLARE outperforms all other methods in both metrics except for the resolution $3\|0$, where SMORE is best.

In Fig.~\ref{fig:cdsc-oasis}, we show the cDSC for the nine regions of interest calculated by SLANT~\cite{huo20193d}. 
Overall we see similar trends as in Fig.~\ref{fig:psnr-ssim-oasis}, with ECLARE performing best at all resolutions except at $3\|0$, where SMORE performs best. 
An exception to this is the cerebellar white matter, where at nearly every resolution SMORE is most performant. 
An interesting observation is how robust SLANT is to interpolation artifacts; bicubic interpolation outperforms single-contrast INR and SynthSR consistently across all resolutions and structures.

\begin{figure*}[!t]
    \centering 
    \includegraphics[width=\linewidth]{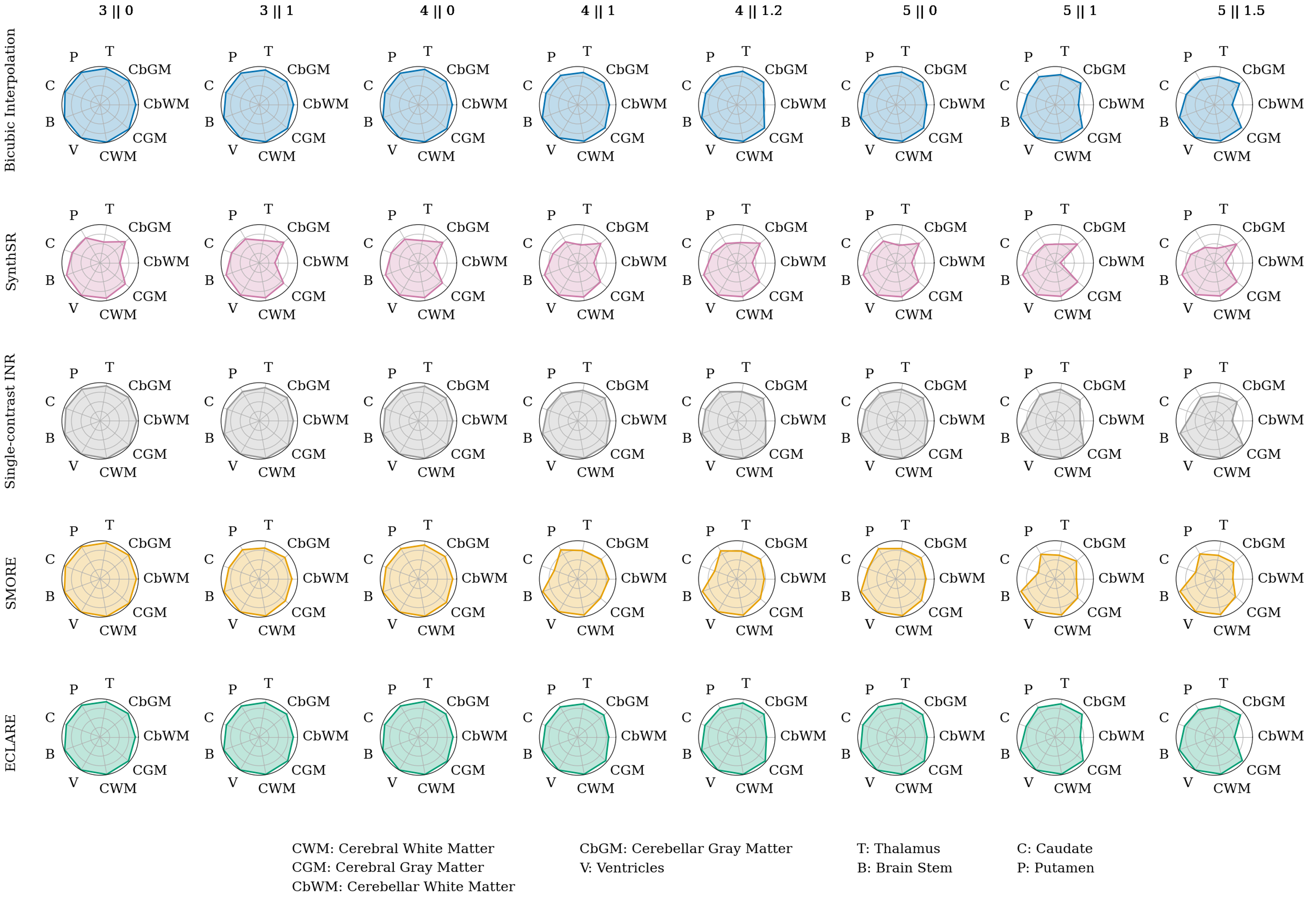}
    \caption{Radar plots showing the $R^2$-value for a linear regression of all $N=50$ subjects between SynthSeg volume estimation on the super-resolution method vs. the SynthSeg volume estimation of the ground truth.}
    \label{fig:radar-synthseg}
\end{figure*}

We also used SynthSeg~\cite{billot2023synthseg} as a downstream task. 
For all subjects, we fit a linear regression model to the sets of estimated volumes against the GT volumes per region per resolution where the dependent variable was the GT volume and the independent variable was the SR volume. 
The fit of the line is quantified by the $R^2$ coefficient of determination, which we plot in Fig.~\ref{fig:radar-synthseg}. 
Some regions (ventricles, cerebral white matter, brainstem) are consistently well-segmented at all resolutions across all SR methods, but others show clear winners. 
ECLARE outperforms all methods consistently, visualized by the enclosure of other methods on the radar plots.

\subsection{People with MS: \texorpdfstring{\TTwo}{T2}-w FLAIR simulations}
\label{sec:3dmrms}
One prominent characteristic of MS is the presence of pathology as white matter hyperintense lesions.
Data-driven SR algorithms conventionally must include relevant pathology in their training data to be expected to perform well on pathological data.
ECLARE, SMORE, and single-contrast INR accomplish this intrinsically since they train on in-plane slices, many of which contain lesions.
For the 3D-MR-MS dataset, we used the automated white matter lesion segmentation algorithm SELF~\cite{zhang2024harmonization} to produce labels for the GT and super-resolved volumes, comparing results with cDSC. 
We also included signal recovery metrics PSNR and SSIM as with the previous dataset.

\begin{figure*}[!t]
    \centering
    \includegraphics[width=\linewidth]{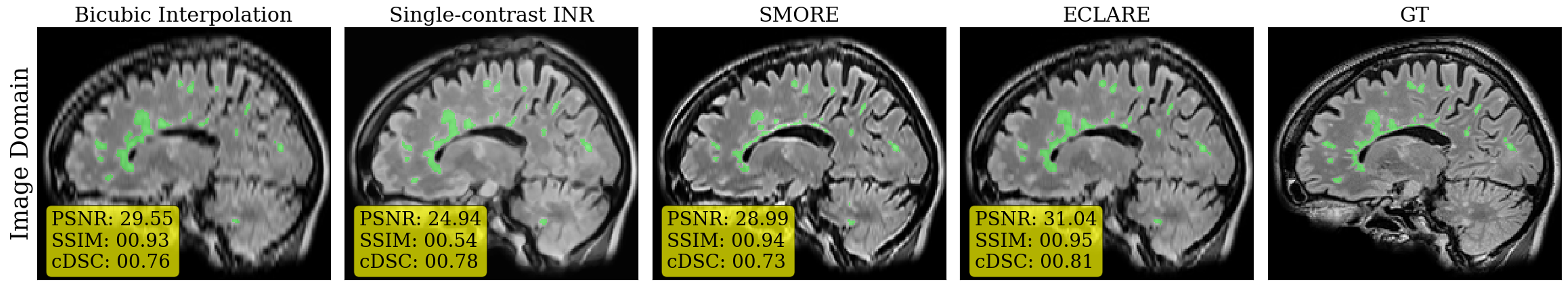}
    \caption{A $4\|1.2$ mid-sagittal slice for a representative subject from the 3D-MR-MS dataset. Each column shows the MR image with the automated lesion segmentation overlaid in neon-green. The PSNR, SSIM, and cDSC for the entire image volume of this representative are overlaid.}
    \label{fig:qual-MS}
\end{figure*}

We performed similar analyses for the 3D-MR-MS dataset, but with SynthSR omitted due to the image being \TTwo-w FLAIR. 
Qualitative results are shown in Fig.~\ref{fig:qual-MS}, where the automated segmentation by SELF is overlaid in neon-green. 
This \TTwo-w FLAIR clearly shows the ``over-sharpening'' artifact from SMORE in the hyperintense region inferior to the ventricle. 
Many such examples of hyperintense regions just inferior to a dark surface are visible in the SMORE result, whereas the ECLARE result does not exhibit this. 
Furthermore, ECLARE better recovers from aliasing artifacts induced by the gap, since this image is $4\|1.2$. 

\begin{figure*}[!t]
    \centering 
    \includegraphics[width=\linewidth]{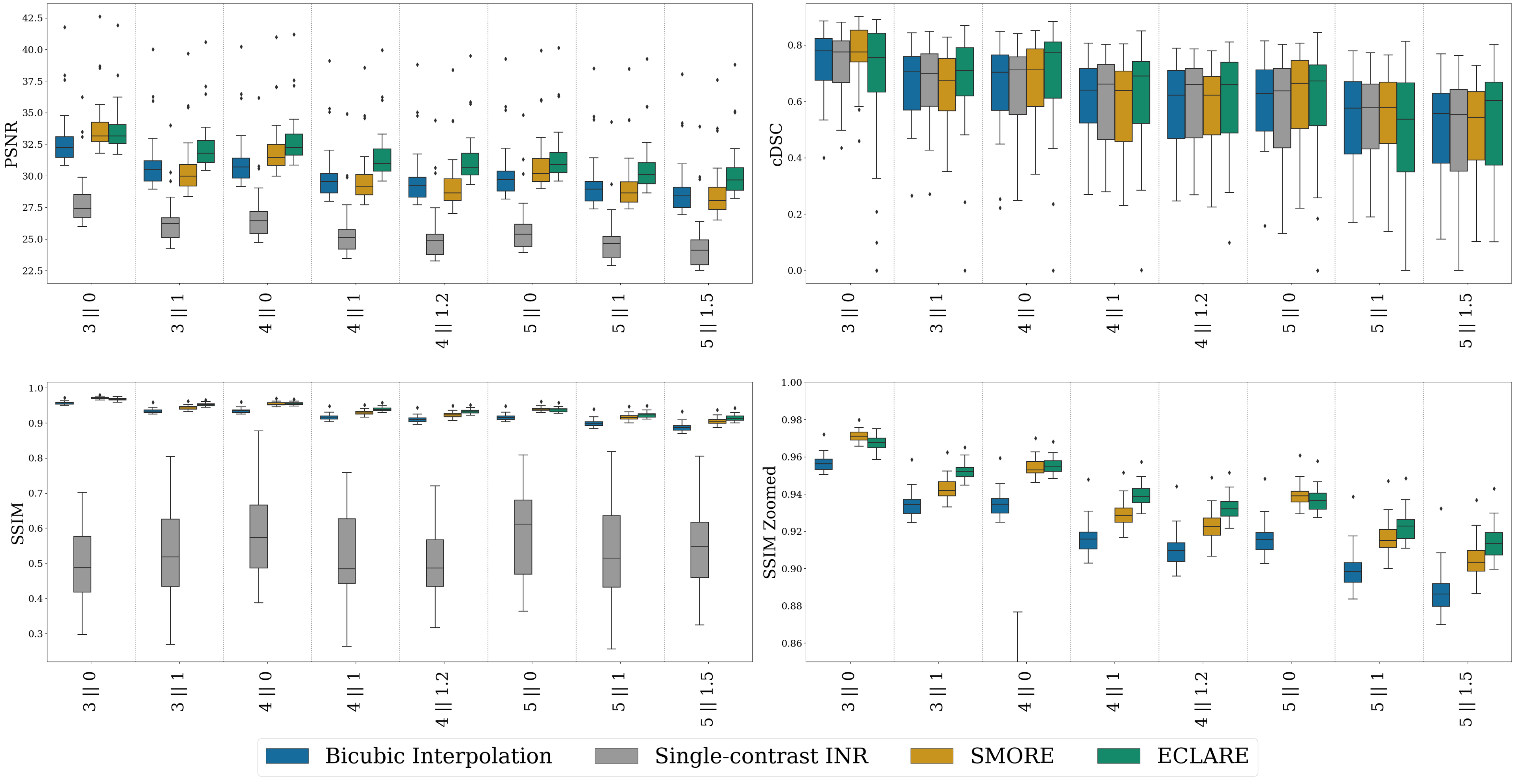}
    \caption{Quantitative results (PSNR in the top-left, cDSC in the top-right, SSIM in the bottom-left, and a zoomed y-axis of SSIM in the bottom-right) on the 3D-MR-MS dataset across resolutions. The bottom-right zoomed SSIM is to better illustrate comparisons between methods with more accurate SSIM scores.}
    \label{fig:quant-MS}
\end{figure*}

In Fig.~\ref{fig:quant-MS}, we show the PSNR and SSIM for signal recovery and cDSC for lesion segmentation consistency. 
Like in the OASIS-3 dataset results, SMORE is best at $3\|0$, though in this case it is not a clear winner. 
The upper quartile appears to outperform ECLARE, but the mean is lower.
These results are consistent across all three metrics. 
At all resolutions worse than $3\|0$, ECLARE has the best performance among all algorithms.

\section{Discussion and conclusions}
\label{sec:discussion}
In this paper we presented ECLARE, a novel self-SR algorithm for anisotropic volumes that trains solely on data extracted from in-plane slices and thus requires no external training data.
We presented a model of the imaging process in 2D multi-slice MRI and used ESPRESO~\cite{han2023espreso} to estimate the slice selection profile.
With an estimated slice profile, slice thickness and slice separation are both modeled, allowing for training data simulation from the HR in-plane data already present in the anisotropic volume.
ECLARE uses a neural network architecture based on WDSR to efficiently train weights from scratch in a short amount of time, taking approximately 5 minutes on a single NVIDIA V100 GPU with a memory footprint under 2~GB.
The result is a SR technique that respects the LR image contrast and presence of pathology, does not require external training data, and produces HR image volumes with the correct digital resolution and FOV.
In the ablation study, we observed that the individual technical contributions of ECLARE are not independently additive, but instead exhibit strong interdependencies.
In particular, ESPRESO alone did not improve quantitative performance and, when used without appropriate resampling, could even degrade PSNR and SSIM.
However, when combined with either FOV-aware resampling or the modified WDSR architecture, ESPRESO consistently led to improved performance, suggesting that accurate modeling of slice selection profiles is most effective when paired with resampling strategies that respect acquisition geometry.
These results indicate that the full ECLARE framework benefits from the complementary interaction of all three components, rather than from any single contribution in isolation.
When compared to four other approaches on two simulated datasets, our approach achieved superior quantitative and qualitative results with respect to signal recovery metrics PSNR and SSIM as well as a downstream task metric---cDSC for brain region and MS lesion segmentation. 
This is true for all investigated resolutions except for $3\|0$, where SMORE performs best.
This is likely due to two reasons: 1)~SMORE does not have a model of slice gap, and therefore will perform best when no gap is present in the data; and 2)~SMORE relies on pre-trained weights and was primarily tested and evaluated using data with resolutions $2\|0$ and $3\|0$. 

We note that computation time is a primary limitation of ECLARE.
ECLARE must fit a model for each multi-slice 2D MR volume.
Although its training time is relatively fast compared to other self-SR methods, it is slower than pre-trained models that perform inference only.
It would be possible to pre-train-then-fine-tune ECLARE as SMORE, but this introduces a tradeoff between time and relying on pre-training as a prior more than the in-plane data itself.
Future work will investigate these tradeoffs in detail.
Additionally, future work will involve experimental design to evaluate the viability of super-resolution in clinical practice.

The SR of medical images has been a research area for decades.
Data-driven approaches now dominate the field of SR, yet the problem is far from solved.
When the test-time image does not belong to the training domain, performance of trained approaches is unpredictable.
In medical imaging, domain issues may occur across a number of axes, including resolution differences, anatomical variations, presence of pathology, differences in contrast or scan acquisition parameters, noise level and distribution mismatch, or others.
Our proposed method ECLARE considers all of these by training on in-plane data and leveraging a forward model inspired by the MR acquisition physics.
Because it does not rely on external data for prior information, ECLARE has the potential to be used on any volumetric image and is not limited to scenarios for which a large amount of data---matched in both contrast and resolution---can be used for training.

\subsection*{Disclosures}
\label{sec:disclosures}
This research is based on technology described in U.S. Patent No.~$12,067,698$, titled \textit{"Machine learning processing of contiguous slice image data,"} assigned to Jerry L. Prince, Can Zhao, and Aaron Carass.

\subsection* {Code, Data, and Materials Availability} 
The code associated with this manuscript is open-source and available at \\ \href{https://www.github.com/sremedios/eclare}{https://www.github.com/sremedios/eclare}.
The code associated with the comparison methods SMORE~\cite{zhao2020smore}, single-contrast INR~\cite{mcginnisandshit2023inr}, and SynthSR~\cite{iglesias2023synthsr}, as well as the code for the two downstream brain parcellation methods SLANT~\cite{huo20193d} and SynthSeg~\cite{billot2023synthseg} are available from their respective online sources.
The data used in this study, OASIS-3~\cite{LaMontagne2019_OASIS3} and 3D-MR-MS~\cite{lesjak2018novel}, are open-source and available from their respective online sources.

\subsection* {Acknowledgments}
This material is supported by the National Science Foundation Graduate Research Fellowship under Grant~No.~DGE-1746891~(Remedios). Development is partially supported by NIH ORIP grant R21~OD030163~(Pham). This work also received support from National Multiple Sclerosis Society RG-1907-34570~(Pham), FG-2008-36966~(Dewey), CDMRP~W81XWH2010912~(Prince), NIH~K01EB032898~(Schilling), and the Department of Defense in the Center for Neuroscience and Regenerative Medicine. The opinions and assertions expressed herein are those of the authors and do not reflect the official policy or position of the Uniformed Services University of the Health Sciences or the Department of Defense.


\bibliography{report}   
\bibliographystyle{spiejour}   





\end{document}